\documentclass[letterpaper, 10 pt, conference]{ieeeconf}  
\pdfoutput=1
\IEEEoverridecommandlockouts                             
\renewcommand{\baselinestretch}{0.97}
\usepackage[T1]{fontenc} 
\usepackage[x11names]{xcolor}
\usepackage{tikz}
\usetikzlibrary{patterns,shapes.arrows, arrows, positioning, backgrounds, fit, calc}
\tikzstyle{block}=[align=center,fill=gray!10,rounded corners,draw=black]
\usepackage{graphicx} 
\graphicspath{{figures/}}
\usepackage[style=ieee, url=false, doi=false, natbib=true, mincitenames=1, maxcitenames=1, maxbibnames=99]{biblatex}
\usepackage[bookmarks=true, hidelinks]{hyperref}
\usepackage{amsmath}
\usepackage{amssymb}
\usepackage{siunitx}
\usepackage{mathtools}
\usepackage[capitalise]{cleveref}
\usepackage{caption}
\usepackage{subcaption}
\usepackage{booktabs} 
\usepackage{flushend}
\usepackage{pgfplots}
\usepgfplotslibrary{groupplots,dateplot}
\pgfplotsset{compat=newest, every tick label/.append style={font=\footnotesize}}
\usepackage{pgfkeys}
    \newenvironment{customlegend}[1][]{%
        \begingroup
        \csname pgfplots@init@cleared@structures\endcsname
        \pgfplotsset{#1}%
    }{%
        \csname pgfplots@createlegend\endcsname
        \endgroup
    }%
    \def\addlegendimage{\csname pgfplots@addlegendimage\endcsname}
\usepackage{stackengine}
\newcommand\solidcirc[4][0]{\rotatebox{#1}{\tikz{\draw[line width=#2] (0,0) 
  arc [x radius=#3,y radius=#4,start angle=0,end angle=360];}}}
  
\title{\LARGE \bf
Predicting Future Spatiotemporal Occupancy Grids \\ with Semantics for Autonomous Driving
}

\author{Maneekwan Toyungyernsub, Esen Yel, Jiachen Li, and Mykel J. Kochenderfer
\thanks{The authors are with the Stanford Intelligent Systems Laboratory (SISL), Stanford University. 
Email: {\tt\footnotesize \{maneekwt, esenyel, jiachen\_li, mykel\}@stanford.edu}.}
}
\begin{document}
\maketitle
\thispagestyle{empty}
\pagestyle{empty}
\begin{abstract}
For autonomous vehicles to proactively plan safe trajectories and make informed decisions, they must be able to predict the future occupancy states of the local environment. 
However, common issues with occupancy prediction include predictions where moving objects vanish or become blurred, particularly at longer time horizons.  
We propose an environment prediction framework that incorporates environment semantics for future occupancy prediction. Our method first semantically segments the environment and uses this information along with the occupancy information to predict the spatiotemporal evolution of the environment. We validate our approach on the real-world Waymo Open Dataset. Compared to baseline methods, our model has higher prediction accuracy and is capable of maintaining moving object appearances in the predictions for longer prediction time horizons. 
\end{abstract}
\section{Introduction}
Accurately predicting the future evolution of the environment around an autonomous vehicle is important to safe navigation.
However, environment prediction is challenging because there are often many different types of road users and the driving environment is crowded. 
In this work, we aim to predict the future spatiotemporal evolution of the local environment represented as occupancy grid maps (OGMs)~\cite{Elfes} by incorporating semantic information of the environment. 

OGMs are widely used in mobile robot mapping and navigation. In the occupancy grid representation, the environment is discretized into cells, where each cell stores a probabilistic estimate of its cell state. OGMs consider the binary free or occupied hypotheses. To incorporate occlusion, another representation called evidential occupancy grid maps (eOGMs)~\cite{Pagac} can be used,
where each cell holds the additional occluded channel for the occluded occupancy hypothesis, along with the free and occupied channels. 

The occupancy grid representations, including both OGMs and eOGMs, are very similar to RGB images in both their spatially discretized forms and information channels. Therefore, the task of predicting future occupancy grids can be posed as a task of predicting future video frames. 
Following the numerous successes in using deep learning for video frame prediction~\cite{PredNet, mim, ruben_prediction, predrnn++}, various video-frame-prediction neural networks have been repurposed to predict the future occupancy states in the environment~\cite{Masha, Hoermann2018DynamicOG, Schreiber, Mohajerin_2019_CVPR, mery_icra, mery_iros, Lange_iros, deep_tracking}.

\citet{Masha} use the PredNet~\cite{PredNet} architecture that is based on a convolutional recurrent neural network to predict the future occupancy states.
The relative motion between the stationary and moving obstacles is accurately captured in the predictions, but the approach suffers from moving object disappearance in the predictions at longer prediction time horizons. Subsequent attempts to prevent vanishing objects 
work by directly incorporating environment dynamics into the model in the form of a double-prong architecture~\cite{mery_icra, mery_iros}, or by augmenting with attention mechanism~\cite{Lange_iros}. 

\begin{figure}[t]
\centering
\input{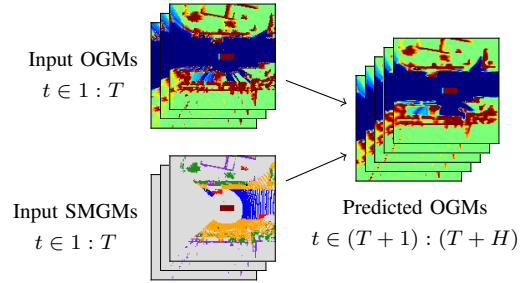}
\caption{\small Future occupancy states are predicted from the past occupancy and semantic inputs.}
\label{fig:teaser}
\end{figure}

\begin{figure*}[t]
    \centering
    \input{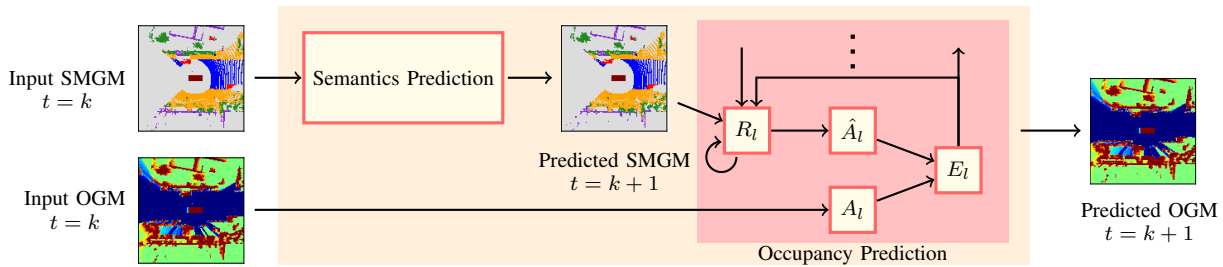}
    \caption{\small 
    The environment prediction model consists of the SMGM and OGM prediction modules. The predicted SMGM output is subsequently used as an input to the representation layer of the occupancy prediction module to predict future occupancy states. The occupancy prediction module is based on the PredNet~\cite{PredNet} architecture and is modified to incorporate semantic information. The figure shows the next-frame prediction ($t+1$). For multiple-frame prediction, the model treats both the previous OGM and SMGM predictions as inputs and recursively iterates to make next-frame predictions.}
    \label{fig:model_illustration}
\end{figure*}

Inspired by the success of incorporating environment dynamics 
directly into the model~\cite{mery_icra, mery_iros}, we develop the future occupancy prediction model that incorporates environment semantics. 
The model receives as input the history of both the past occupancy information (OGMs), as well as the past semantic information in the form of semantic grid maps (SMGMs), and predicts the future occupancy states in the environment, as illustrated in \cref{fig:teaser}.
With our proposed SMGMs, the model has access to semantic information on the different types of agents present in the scene, e.g., vehicles, cyclists, and pedestrians. In comparison, the double-prong model only perceives the objects as belonging to either static or moving categories in prior work~\cite{mery_iros}. We hypothesize that, with the addition of semantics, our proposed model can better learn and capture different motion models for dynamic objects belonging to different categories (e.g., between vehicles and slower-moving cyclists or pedestrians), leading to an improvement in occupancy state prediction.       

Our contributions are as follows. 
We propose to represent environment semantics in the form of SMGMs, and develop a method that incorporates the semantic information within the occupancy prediction framework. 
The proposed model internally 
predicts future environment semantics and passes the 
future semantic information directly to the occupancy prediction module, as illustrated in~\cref{fig:model_illustration}. Our approach is validated on the real-world Waymo Open Dataset (v1.4.0)~\cite{Sun_2020_CVPR} and achieves better performance than other state-of-the-art methods in future occupancy prediction. 

\section{Related Work}

\subsection{Predicting Future Occupancy States}
Successes in deep learning have led to the development of several neural network-based occupancy prediction methods~\cite{Masha, Hoermann2018DynamicOG, Schreiber, Mohajerin_2019_CVPR, mery_icra, Lange_iros, mery_iros, deep_tracking} using variants of convolutional neural network (CNN) and recurrent neural network (RNN) architectures.  
\citet{Hoermann2018DynamicOG} present a method based on a deep CNN architecture to predict the future states in the environment represented in the form of dynamic occupancy grid maps (DOGMas)~\cite{Nuss_DOGMa}.
Subsequent work by \citet{Schreiber} introduces a prediction framework, based on a convolutional long short-term memory architecture~\cite{convlstm}, and achieves better results on future DOGMa prediction. 
In addition to the occupancy probability, the DOGMa representation carries the cell-wise velocity estimate obtained from a particle filter, which is very computationally expensive.

Vanishing moving objects in the predictions, especially at longer prediction time horizons, are common occurrences~\cite{Masha, mery_icra, mery_iros, Lange_iros, li2020evolvegraph, choi2021shared}. 
A recent attempt to counter the disappearance problem works by directly integrating dynamic information of the environment into a double-prong model, based on the PredNet architecture~\cite{PredNet}, by means of separating the OGM inputs into globally static OGMs and dynamic OGMs, and feeding these into the static, and dynamic prong, 
respectively~\cite{mery_icra, mery_iros}.
Subsequent work by~\citet{mery_iros} integrates static-dynamic object segmentation module into their method to determine stationary and moving objects in the environment, instead of relying on object detection and tracking information as done in the prior work~\cite{mery_icra}. \citet{lange2023lopr} reduce blurriness in predictions by making predictions in the latent space of a generative model.
Different from these works, we propose using environment semantics to retain valuable contextual information in the scene, rather than only giving the model access to static and moving object information.      

Previous works by~\citet{Mann} and \citet{asghar2023vehicle} present occupancy grid prediction models that use semantic labels of the occupied cells to improve prediction.~\citet{Mann} introduces a model that predicts occupancy grids of the vehicles and the environment separately. Thresholding is used in their approach to convert the OGMs (containing occupancy probability) into binary OGMs to represent occupied and unoccupied states.~\citet{asghar2023vehicle} present a framework that predicts future occupancy states represented as dynamic occupancy grid maps (DOGMas) and requires prior map information as an additional input to the architecture. The DOGMa representation requires a particle filter to estimate the cell-wise velocities, which are computationally expensive. In these prior work, the semantic labels consist of only the vehicle class, whereas our proposed SMGMs contain other semantic classes in addition to the vehicle label (e.g., cyclists and pedestrians) to encourage the model to learn different motion models between different dynamic entities.

\subsection{Predicting Future Semantic Segmentation Maps}

While the video frame prediction task (predicting raw RGB pixel values) has been extensively studied~\cite{PredNet, mim, ruben_prediction, predrnn++}, the task of predicting semantic segmentation maps of future video frames was introduced recently in the study by~\citet{Luc2017PredictingDI}. 
Their method is based on an autoregressive CNN architecture that learns to predict multiple future frames given a sequence history of past frames.
Their experiments show that directly making predictions 
in the semantic space
is more beneficial than first predicting future RGB images and then performing semantic segmentation on them. They conclude that their neural network architecture learns the interaction dynamics and simple physics better when modeling pixel-wise object classes instead of RGB values.
Inspired by their findings, we modify the PredNet architecture~\cite{PredNet} designed for video frame prediction to predict the future environment semantics in the form of semantic grid maps (SMGMs) instead.
Our aim is to provide the occupancy prediction model with the encoded future semantic information via the predicted SMGMs, thereby allowing the network to better model the dynamics between different objects.

\section{Approach}

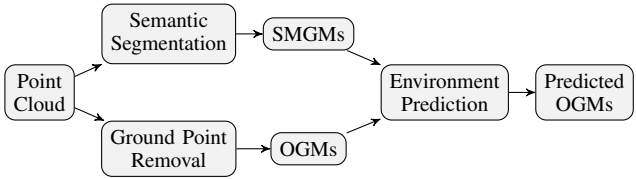
\begin{figure}
    \centering
    \begin{tikzpicture}[>=stealth',font=\footnotesize]
\matrix [column sep=0.35cm,row sep=0.001cm] {
& \node[block] (sem_seg) {Semantic \\ Segmentation}; & 
\node[block] (smgm) {SMGMs}; \\
\node[block] (pc) {Point \\ Cloud}; & & &
\node[block] (env_pred) {Environment \\ Prediction}; &
\node[block] (pred_ogm) {Predicted \\ OGMs}; \\
& \node[block] (ground_seg) {Ground Point \\ Removal}; & 
\node[block] (ogm) {OGMs}; \\
};
\draw [->] 
    (sem_seg) edge (smgm)
    (pc) edge (sem_seg)
    (pc) edge (ground_seg)
    (ground_seg) edge (ogm)
    (smgm) edge (env_pred)
    (ogm) edge (env_pred)
    (env_pred) edge (pred_ogm)
;
\end{tikzpicture}
    \vspace{-1.5em}
    \caption{\small The whole pipeline for our proposed methodology. Semantic grid maps (SMGMs) are generated from the semantic annotations predicted from the semantic segmentation module. The environment prediction model outputs the future OGM predictions from the past OGM and SMGM inputs.}
    \vspace{-2em}
    \label{fig:pipeline}
\end{figure}

Our goal is to develop an environment prediction framework that incorporates scene semantics within the occupancy prediction model, as illustrated in \cref{fig:model_illustration}.
We consider dividing the model into two modules as part of the environment prediction. The upstream semantics prediction module determines how the environment semantics evolves temporally. 
The downstream occupancy prediction module incorporates the semantic information provided by the upstream module, and subsequently predicts the future occupancy states in the environment. 
The whole pipeline for the proposed framework is shown in \cref{fig:pipeline}. 
The environment representation, in the form of OGMs, is formed using data from range-bearing sensors after first implementing removal of ground points using a Markov Random Field~\cite{Postica}, as performed by~\citet{Masha}. In order to form the semantic grid maps (SMGMs), we first perform semantic segmentation to obtain the semantic annotations associated with each data point. 
\subsection{Occupancy Representation}
Following prior work on spatiotemporal occupancy prediction~\cite{Masha, mery_icra}, we choose to represent the environment around the ego vehicle in the form of eOGMs~\cite{Pagac}, which is an alternative probabilistic representation of OGMs. 
The environment is discretized into grid cells, where each cell contains the probabilistic estimate of its respective belief of occupancy. The eOGM representation is formed using Dempster--Shafer Theory (DST)~\cite{dst} to update the occupancy belief of each cell. 
Since eOGMs consider each cell to be either occupied or empty, the allowable hypotheses are occupied, empty, or unknown, $\{\{O\}, \{E\}, \{O, E\} \}$. The measure of belief of occupancy for each cell is represented by the belief masses associated with each allowable hypothesis. The eOGMs are in $\mathbb{R}^{W \times H \times C}$, where $W$, $H$, and $C$ are the width, height, and the number of channels.
Since the belief masses of each allowable hypothesis should sum to $1$, we consider the belief masses for the empty ($m(\{E\}) \in [0, 1]$) and occupied ($m(\{O\}) \in [0, 1]$) hypotheses for each cell.
Thus, the eOGMs are represented with two channels. For visualization purposes, the occupied and empty belief masses in the eOGM representation can be converted to an estimate of the occupancy probability (OGMs) using the pignistic transformation~\cite{pignistic}, as done in~\cite{Masha}, and defined as follows,
\begin{align}
\label{eq:P_O}
p(O) = 0.5 \times (1-m(\{E\})) + 0.5 \times m(\{O\}).
\end{align}

\subsection{Semantic Representation}
\label{section:approach:semantic}
Inspired by the work on spatiotemporal occupancy prediction using environment dynamics 
~\cite{mery_icra, mery_iros}, we propose incorporating environment semantics in the occupancy prediction model.
Our hypothesis is that an improvement in the prediction accuracy could be achieved if the model also has access to the cell-wise object label information in addition to the occupancy state information. 
Our aim is to provide cues to the model when predicting the occupancy states for different objects, including static objects and different movable agents. The model should learn to capture and translate the relative motion of the ego vehicle when predicting the occupancy states of static objects and learn to predict different motion models between different types of movable agents (e.g., pedestrians and faster-moving vehicles).  

To incorporate the semantic information, we propose representing the environment semantics in the form of SMGMs. The SMGMs are in $\mathbb{R}^{W \times H}$, representing the same local environment space around the ego vehicle as the OGMs, and carry the cell-wise object label information. The SMGMs are constructed in a similar way to the OGMs, by assigning the cell corresponding to a sensor measurement with its respective object label.

To obtain the required object annotations, we use the SalsaNext~\cite{salsanext} architecture to perform semantic segmentation on the point cloud data before generating the SMGMs, instead of relying on the ground truth semantic data. We consider semantic segmentation since, in practice, the model does not have access to ground truth semantics a priori.  
It is noted that the model performance therefore relies on the accuracy of the upstream semantic segmentation module.
We choose to use the SalsaNext network due to its good segmentation performance as shown in prior work~\cite{salsanext, mos}. 

\subsection{Environment Prediction Model}
To incorporate semantics, we design the environment prediction model to consist of two modules, as illustrated in \cref{fig:model_illustration}. The upstream semantics prediction module receives SMGMs as input and learns to predict the spatiotemporal evolution of the environment semantics. The downstream occupancy prediction module then receives the predicted future SMGMs, along with a history of past OGMs, and learns to predict the future occupancy states of the environment.

\subsubsection{Occupancy Grid Prediction}
Following prior occupancy prediction work~\cite{Masha, mery_icra}, we adapt the video-frame prediction PredNet~\cite{PredNet} architecture as our occupancy prediction module, with some modifications to the network to allow for the flow of semantic information. The original PredNet architecture consists of a series of stacked modules, where the representation layer $R_l$ of each module is trained to make predictions $\hat{A_l}$ of the module-specific input $A_l$, and the error $E_l$ between the prediction and the target is propagated within the same module, and also the next module in the series. 

Our proposed modification is in the recurrent representation layer, where we connect the information flow for environment semantics from the output layer of the upstream semantics prediction module to the first representation layer of the occupancy prediction module, as shown in \cref{fig:model_illustration}. By connecting the information flow in this manner, we avoid the need to feed the predicted SMGMs from the upstream module to the input layer $A_l$, along with the input OGMs. Since $A_l$ is the prediction target, feeding SMGMs here otherwise requires the module to learn to predict both the future semantics and occupancy states simultaneously. We are interested in accurate predictions of future OGMs, while using semantics to provide additional cues when making the prediction. By separating the semantic and occupancy modules and connecting the semantic information flow through the representation layer, we can train the occupancy prediction module to focus solely on predicting future OGMs. 

\subsubsection{Semantic Grid Prediction}
Inspired by the successful repurposing of the video-frame prediction PredNet~\cite{PredNet} architecture for the spatiotemporal occupancy prediction task~\cite{Masha, mery_icra}, we adapt the PredNet architecture to predict the evolution of scene semantics with minor modifications to the network. Since the task of predicting future scene semantics can be framed as a multi-class classification task, we use the softmax activation instead of the original rectified linear unit (ReLU) activation in the output prediction layer. Additionally, the loss function is changed to the categorical cross-entropy loss to suit the classification problem instead.

\section{Experiments}
\subsection{SMGM and OGM Generation}
\label{section:exp}
The SMGM and OGM data are generated from the latest version of the  Waymo Open Dataset (v1.4.0)~\cite{Sun_2020_CVPR}, which has been expanded to include $3$D semantic segmentation labels. However, the provided annotations exist for certain selected frames and comprise about $15$\% of the entire dataset. After removing points belonging to the ground from the LiDAR data, the OGMs are generated with a width and a height of $128$ cells, with a cell resolution of \SI{0.33}{\metre}. This is equivalent to an approximate area of $\SI{42}{\metre} \times \SI{42}{\metre}$. 
These dimensions are chosen to match with the prior work~\cite{Masha, mery_iros} so comparisons can be made.
The OGMs represent the relative motion of the ego vehicle with the environment since the ego vehicle is fixed to the center of every frame.  
For the SMGMs, we first perform semantic segmentation, using the provided semantic segmentation labels as ground truths to train the model. The resultant model is then used to generate the semantic annotations for the entire dataset, and we use these predictions to subsequently generate the SMGMs, without relying on the semantic ground truths. SMGMs are generated to match the width and height of the OGM data. 
Note that semantic segmentation is performed on the entire point cloud, including the ground points so that we can provide the ground object label to the prediction model. The semantic labels consist of car, other vehicles (e.g., train and bus), bicyclist, pedestrian, traffic object, building, vegetation, road, undrivable surface (e.g., curb and sidewalk), ego vehicle, and \emph{others} (containing miscellaneous objects). 

\subsection{Semantics and Occupancy Model Training}

Our proposed environment prediction model receives as input a history of past OGMs and SMGMs, and the output is the future OGM predictions of the local environment around the ego vehicle. 
Both the OGMs and SMGMs are arranged into sequences, each consisting of $20$ frames and is equivalent to \SI{2}{\second} of data per sequence.  
For a given sequence, the model predicts the future $15$ OGMs, based on the past $5$ OGMs and SMGMs, as done in \cite{Masha, mery_iros} for performance comparison later. 
The train, validation, and test split ratios are $0.7$, $0.15$, and $0.15$, respectively.
The Adam optimizer~\cite{adam} is used during training with a starting learning rate of $0.0001$ for $80$ epochs with a batch size of $16$.
Since the environment prediction model is designed such that the semantic and the occupancy prediction modules are two separate entities, these two models can, therefore, be trained independently of each other. The semantics prediction module is trained first, and its weight and bias parameters are kept frozen while training the occupancy prediction module. As recommended by~\citet{PredNet}, for multiple-frame prediction tasks, the model is trained in $2$ stages. In the first stage, the model is trained to predict the next-frame OGM, given the current OGM and SMGM as an input. 
Using the weight and bias parameters from the first stage for weight initialization, we then fine-tune the model to make multiple-frame predictions in the second training stage.
The model is trained to treat the last-frame prediction as an actual input and predict the next OGM frame in a recursive manner for all $15$ prediction time steps.
The inference time is \SI{130}{\milli\second} per sample on an Intel i7-5930K at 3.5 GHz and an NVIDIA GeForce TITAN X graphics card.

\subsection{Performance Evaluation}

We compare our prediction results against prior occupancy prediction work.
The first baseline uses only occupancy information to predict the future OGMs~\cite{Masha}. The second baseline integrates environment dynamics into occupancy prediction by first segmenting the environment into static and dynamic parts through its static-dynamic object segmentation module. The OGMs are then split into static and dynamic OGMs as input to the double-prong model for future OGM predictions~\cite{mery_iros}. To evaluate performance, we use the mean squared error (MSE) and the image similarity (IS)~\cite{im_sim} metrics between the ground truth labels and predicted OGMs. The MSE metric is used as a measure of how accurately the predicted value of occupancy probability corresponds to its target value. 
The IS metric is a measure of how well the predictions retain the structure of the environment. 
To obtain the IS value, the minimum Manhattan distance is computed between the two closest cells having the same occupancy class (from occupied, empty, and unknown classes) between the target OGMs and the predictions. 

\section{Results}
\begin{figure*}[t!]
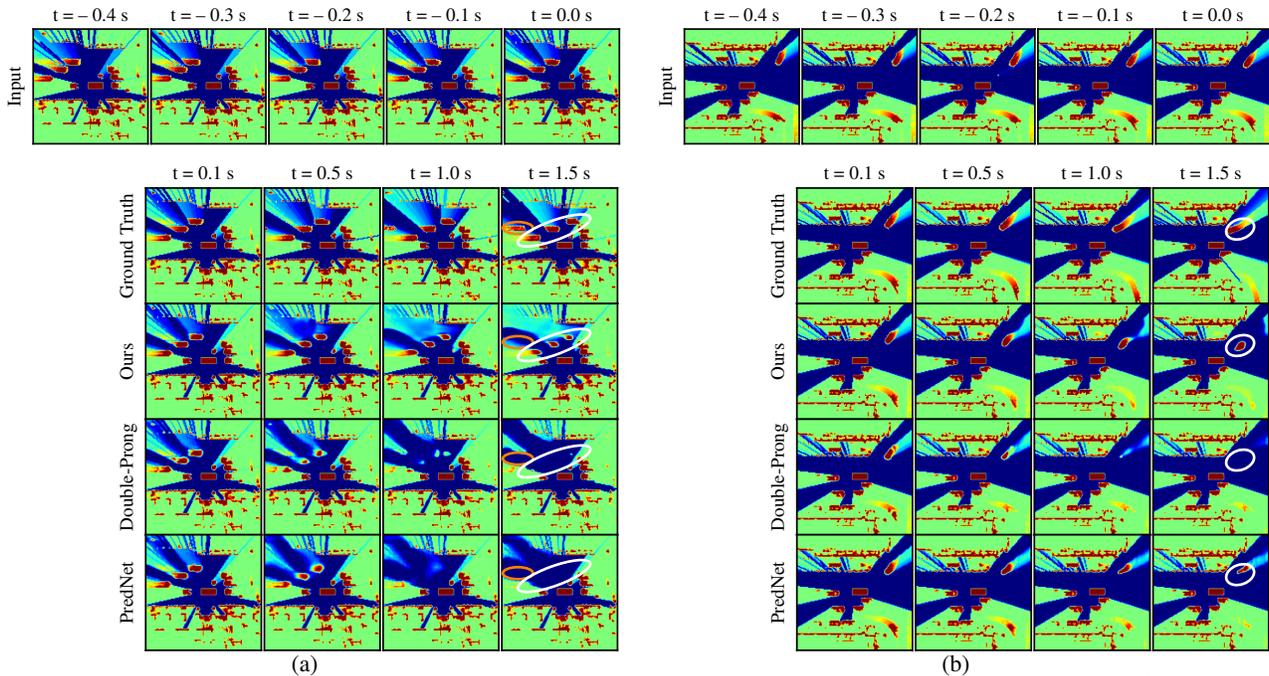

\centering
\begin{subfigure}[t]{0.48\textwidth}
\centering
\stackinset{l}{2.77in}{b}{2.2in}{\textcolor{white}{\solidcirc[20]{1pt}{0.51}{0.13}}} 
{\stackinset{l}{2.77in}{b}{1.6in}{\textcolor{white}{\solidcirc[20]{1pt}{0.51}{0.13}}}
{\stackinset{l}{2.77in}{b}{0.99in}{\textcolor{white}{\solidcirc[20]{1pt}{0.51}{0.13}}}
{\stackinset{l}{2.77in}{b}{0.39in}{\textcolor{white}{\solidcirc[20]{1pt}{0.51}{0.13}}}
{\stackinset{l}{2.71in}{b}{2.3in}{\textcolor{orange}{\solidcirc[0]{1pt}{0.2}{0.08}}} 
{\stackinset{l}{2.71in}{b}{1.702in}{\textcolor{orange}{\solidcirc[0]{1pt}{0.2}{0.08}}}
{\stackinset{l}{2.71in}{b}{1.092in}{\textcolor{orange}{\solidcirc[0]{1pt}{0.2}{0.08}}}
{\stackinset{l}{2.71in}{b}{0.492in}{\textcolor{orange}{\solidcirc[0]{1pt}{0.2}{0.08}}}
{\input{Figures/Fig_qualitative/plot_728}}}}}}}}}
\vspace{-0.9cm}
\caption{}
\label{fig:qualitative_straight}
\end{subfigure}
\begin{subfigure}[t]{0.48\textwidth}
\centering
\stackinset{l}{3.07in}{b}{2.25in}{\textcolor{white}{\solidcirc[20]{1pt}{0.192}{0.13}}} 
{\stackinset{l}{3.07in}{b}{1.64in}{\textcolor{white}{\solidcirc[20]{1pt}{0.192}{0.13}}}
{\stackinset{l}{3.07in}{b}{1.04in}{\textcolor{white}{\solidcirc[20]{1pt}{0.192}{0.13}}}
{\stackinset{l}{3.07in}{b}{0.44in}{\textcolor{white}{\solidcirc[20]{1pt}{0.192}{0.13}}}
{\input{Figures/Fig_qualitative/plot_575}}}}}
\vspace{-0.9cm}
\caption{}
\label{fig:qualitative_turning}
\end{subfigure}
\vspace{-0.3cm}
\caption{\small Example driving scenarios and their OGM predictions (occupied: red, unknown: green, empty: blue). The predicted OGMs are shown at selected prediction time steps of \SI{0.1}{\second}, \SI{0.5}{\second}, \SI{1.0}{\second}, and \SI{1.5}{\second}. \cref{fig:qualitative_straight} is a scenario with multiple vehicles traveling straight. \cref{fig:qualitative_turning} is a scenario with a vehicle making a right turn. Compared to baseline methods, our model maintains the predicted moving object appearances for longer prediction time horizons in both example scenarios.}
\vspace{-0.6cm}
\label{fig:qualitative}
\end{figure*}
We evaluate our approach against baseline methods to assess the performance gains of incorporating environment semantics into the occupancy prediction model (ours) against one that has no access to semantic information (PredNet~\cite{PredNet}). Additionally, we also compare against a model that incorporates dynamics (the double-prong model~\cite{mery_iros}) to assess the the effectiveness between incorporating environment semantics against environment dynamics. 

\subsection{Qualitative Evaluation}

\cref{fig:qualitative} illustrates two example driving scenarios and their resultant OGM predictions. 
The inputs to all models are the past $5$ OGM frames, as shown on the top row in \cref{fig:qualitative}, and the past $5$ SMGM frames for our method. The ground truth and predicted OGMs are depicted at selected prediction time steps of \SI{0.1}{\second}, \SI{0.5}{\second}, \SI{1.0}{\second}, and \SI{1.5}{\second}. Occupied, empty, and unknown cells are depicted in red, blue, and green, respectively. 
The motion of other objects in the environment is shown as relative to the ego vehicle since the ego vehicle is fixed at the center of every frame and is headed to the right.
As \cref{fig:qualitative} illustrates, our model is able to retain moving objects in the predictions for longer prediction time horizons and reasonably capture their positions for a variety of scenes, including when the vehicles are traveling straight (\cref{fig:qualitative_straight}), or when the vehicles are turning (\cref{fig:qualitative_turning}). As shown in \cref{fig:qualitative_straight}, the three moving vehicles (circled in white) are retained in our OGM predictions at all time steps, whereas they start disappearing earlier on in the baseline predictions. They either completely disappear from the predictions (PredNet) or are predicted as being unknown (double-prong) at the \SI{1.0}{\second} prediction time step. Although our model is able to retain these vehicles, their predictions in the OGMs at later prediction time horizons start to appear smaller. This degradation in the prediction accuracy for all models will be described in more detail in \cref{section:quant}. 

For objects that only start to appear in the prediction time horizon, and are not part of the environment in the input OGM history, none of the models are able to predict that these objects exist, as expected. An example of this scenario is shown in \cref{fig:qualitative_straight}, where there is a vehicle (circled in orange) that only starts to appear sometime between the \SI{0.5}{\second} and \SI{1.0}{\second} prediction time steps, and is not a part of the input OGM history. None of the methods contain this moving vehicle in their predictions. 

Predicting the OGMs for fast-moving, or turning vehicles, can be particularly challenging for all models, as illustrated in \cref{fig:qualitative_turning}, where there is a vehicle turning right at an intersection (circled in white). 
Our model is able to maintain the appearance of the turning vehicle at all prediction time steps, although the turning orientation is not captured quite as accurately. The model seems to predict that the vehicle would keep traveling straight ahead instead of making a right turn. 
We hypothesize that the straight-motion prediction could be due to the vehicle seemingly appears to move in a straight direction in the input OGM history. 
In contrast, for the baselines, the predicted turning vehicle either vanishes completely in the predictions or lags behind in position. One possible method to allow the turning motion to be predicted includes modeling the multi-modality in the predictions, which is not within the scope of this study. 
\begin{table*}[t!]
    \centering
    \caption{\small The MSE, IS, and dynamic MSE metrics on predictions averaged over the entire prediction time horizon. Lower is better.}
    \begin{tabular}{@{}lrrrr@{}}
        \toprule
        \multicolumn{1}{c}{\bf Models} &
        \multicolumn{1}{c}{\bf MSE $\times 10^{-2}$} &
        \multicolumn{1}{c}{\bf IS} &
        \multicolumn{1}{c}{\bf Dynamic MSE$\times 10^{-3}$}\\
        \midrule
        Ours (Occupancy + Semantics) & $\mathbf{2.87 \pm 0.0010}$ & $\mathbf{5.05 \pm 0.049}$ & $\mathbf{1.72 \pm 0.0021}$\\
        Double-Prong(Occupancy + Dynamics)~\cite{mery_iros} & $3.52 \pm 0.0011$ & $6.52 \pm 0.059$ & $2.23 \pm 0.0024$\\
        PredNet(Occupancy)~\cite{PredNet} & $3.59 \pm 0.0011$ & $7.44 \pm 0.073$ & $2.12 \pm 0.0024$\\
        \bottomrule
    \end{tabular}
    \label{table:metrics}
\vspace{-0.2cm}
\end{table*}

\begin{figure*}[t!]
    \centering
    \begin{subfigure}[h]{0.28\textwidth}
        \centering
        \begin{tikzpicture}

\definecolor{color0}{rgb}{0.12156862745098,0.466666666666667,0.705882352941177}
\definecolor{color1}{rgb}{1,0.498039215686275,0.0549019607843137}
\definecolor{color2}{rgb}{0.172549019607843,0.627450980392157,0.172549019607843}
\definecolor{color3}{rgb}{0.83921568627451,0.152941176470588,0.156862745098039}

\begin{axis}[
width = 5.5cm,
tick align=outside,
tick pos=left,
x grid style={white!69.0196078431373!black},
xlabel={\footnotesize Prediction Time (s)},
xmin=0.1, xmax=1.5,
xtick style={color=black},
y grid style={white!69.0196078431373!black},
ylabel={\footnotesize MSE},
ymin=0, ymax=0.0576937549747527,
ytick style={color=black}
]
\path [fill=color0, fill opacity=0.2]
(axis cs:0.1,0.010808680133)
--(axis cs:0.1,0.010761860867)
--(axis cs:0.2,0.01667299995)
--(axis cs:0.3,0.01908797478)
--(axis cs:0.4,0.02357405656)
--(axis cs:0.5,0.02672472243)
--(axis cs:0.6,0.03060226446)
--(axis cs:0.7,0.03372629748)
--(axis cs:0.8,0.037086746977)
--(axis cs:0.9,0.040001983646)
--(axis cs:1.0,0.043105910715)
--(axis cs:1.1,0.045841933704)
--(axis cs:1.2,0.04865960613)
--(axis cs:1.3,0.051302757618)
--(axis cs:1.4,0.053847441312)
--(axis cs:1.5,0.056338945673)
--(axis cs:1.5,0.056439702327)
--(axis cs:1.5,0.056439702327)
--(axis cs:1.4,0.053946254687999996)
--(axis cs:1.3,0.051399522382000004)
--(axis cs:1.2,0.04875416387)
--(axis cs:1.1,0.045934034296)
--(axis cs:1.0,0.043195521284999996)
--(axis cs:0.9,0.040088644353999996)
--(axis cs:0.8,0.037170493023000004)
--(axis cs:0.7,0.033806502520000004)
--(axis cs:0.6,0.03067897354)
--(axis cs:0.5,0.02679677757)
--(axis cs:0.4,0.02364200944)
--(axis cs:0.3,0.019149559219999997)
--(axis cs:0.2,0.016730700050000002)
--(axis cs:0.1,0.010808680133)
--cycle;

\path [fill=color2, fill opacity=0.2]
(axis cs:0.1,0.010273157274)
--(axis cs:0.1,0.010227376726000001)
--(axis cs:0.2,0.016353442078)
--(axis cs:0.3,0.019042787285)
--(axis cs:0.4,0.023432667516)
--(axis cs:0.5,0.026656035968)
--(axis cs:0.6,0.030295912919999998)
--(axis cs:0.7,0.033416815770000004)
--(axis cs:0.8,0.036640058796)
--(axis cs:0.9,0.039477889998000004)
--(axis cs:1.0,0.042400081937)
--(axis cs:1.1,0.0450154007)
--(axis cs:1.2,0.047685394047)
--(axis cs:1.3,0.050170773305000006)
--(axis cs:1.4,0.052489539769999996)
--(axis cs:1.5,0.05481927312)
--(axis cs:1.5,0.05491892088)
--(axis cs:1.5,0.05491892088)
--(axis cs:1.4,0.05258734623)
--(axis cs:1.3,0.050266686695)
--(axis cs:1.2,0.047779193953)
--(axis cs:1.1,0.045106845300000004)
--(axis cs:1.0,0.042489118063)
--(axis cs:0.9,0.039564130002)
--(axis cs:0.8,0.036723441204)
--(axis cs:0.7,0.03349678823)
--(axis cs:0.6,0.03037236708)
--(axis cs:0.5,0.026728126031999998)
--(axis cs:0.4,0.023500536484)
--(axis cs:0.3,0.019104438714999998)
--(axis cs:0.2,0.016410717922)
--(axis cs:0.1,0.010273157274)
--cycle;

\path [fill=color3, fill opacity=0.2]
(axis cs:0.1,0.007313990369)
--(axis cs:0.1,0.007274669631)
--(axis cs:0.2,0.011756569831)
--(axis cs:0.3,0.014838590435)
--(axis cs:0.4,0.017858027335999997)
--(axis cs:0.5,0.020905144956000002)
--(axis cs:0.6,0.02375767614)
--(axis cs:0.7,0.02662119689)
--(axis cs:0.8,0.029377468858)
--(axis cs:0.9,0.03205098615)
--(axis cs:1.0,0.03472928743)
--(axis cs:1.1,0.03726774516)
--(axis cs:1.2,0.03975853346)
--(axis cs:1.3,0.042214129722000004)
--(axis cs:1.4,0.04455760785)
--(axis cs:1.5,0.046916319253)
--(axis cs:1.5,0.047009826747)
--(axis cs:1.5,0.047009826747)
--(axis cs:1.4,0.04464904215)
--(axis cs:1.3,0.042303430278)
--(axis cs:1.2,0.03984550654)
--(axis cs:1.1,0.037352262840000004)
--(axis cs:1.0,0.034811182569999995)
--(axis cs:0.9,0.03212998385)
--(axis cs:0.8,0.029453411142)
--(axis cs:0.7,0.02669380511)
--(axis cs:0.6,0.023826573860000002)
--(axis cs:0.5,0.020970079044)
--(axis cs:0.4,0.017918312664)
--(axis cs:0.3,0.014893875564999999)
--(axis cs:0.2,0.011805924169)
--(axis cs:0.1,0.007313990369)
--cycle;
\addplot [semithick, color0, mark=diamond*, mark size=2, mark options={solid}]
table {%
0.1 0.0107852705
0.2 0.01670185
0.3 0.019118767
0.4 0.023608033
0.5 0.02676075
0.6 0.030640619
0.7 0.0337664
0.8 0.03712862
0.9 0.040045314
1.0 0.043150716
1.1 0.045887984
1.2 0.048706885
1.3 0.05135114
1.4 0.053896848
1.5 0.056389324
};

\addplot [semithick, color2, mark=triangle*, mark size=2, mark options={solid}]
table {%
0.1 0.010250267
0.2 0.01638208
0.3 0.019073613
0.4 0.023466602
0.5 0.026692081
0.6 0.03033414
0.7 0.033456802
0.8 0.03668175
0.9 0.03952101
1.0 0.0424446
1.1 0.045061123
1.2 0.047732294
1.3 0.05021873
1.4 0.052538443
1.5 0.054869097
};
\addplot [semithick, color3, mark=*, mark size=2, mark options={solid}]
table {%
0.1 0.00729433
0.2 0.011781247
0.3 0.014866233
0.4 0.01788817
0.5 0.020937612
0.6 0.023792125
0.7 0.026657501
0.8 0.02941544
0.9 0.032090485
1.0 0.034770235
1.1 0.037310004
1.2 0.03980202
1.3 0.04225878
1.4 0.044603325
1.5 0.046963073
};

\end{axis}

\end{tikzpicture}
        \label{fig:mse}
    \end{subfigure}
    \begin{subfigure}[h]{0.28\textwidth}
        \centering
        \begin{tikzpicture}

\definecolor{color0}{rgb}{0.12156862745098,0.466666666666667,0.705882352941177}
\definecolor{color1}{rgb}{1,0.498039215686275,0.0549019607843137}
\definecolor{color2}{rgb}{0.172549019607843,0.627450980392157,0.172549019607843}
\definecolor{color3}{rgb}{0.83921568627451,0.152941176470588,0.156862745098039}

\begin{axis}[
width=5.5cm,
tick align=outside,
tick pos=left,
x grid style={white!69.0196078431373!black},
xlabel={\footnotesize Prediction Time (s)},
xmin=0.1, xmax=1.5,
xtick style={color=black},
y grid style={white!69.0196078431373!black},
ylabel={\footnotesize IS},
ymin=0, ymax=15.2096629642245,
ytick style={color=black}
]
\path [fill=color0, fill opacity=0.2]
(axis cs:0.1,1.0578985630633322)
--(axis cs:0.1,1.00408593360399)
--(axis cs:0.2,2.1731530664269982)
--(axis cs:0.3,2.5078508903516274)
--(axis cs:0.4,3.605906193397436)
--(axis cs:0.5,4.372154967360154)
--(axis cs:0.6,5.382299542674)
--(axis cs:0.7,6.260564003901839)
--(axis cs:0.8,7.171361489486647)
--(axis cs:0.9,8.082579789788609)
--(axis cs:1.0,9.006873322322805)
--(axis cs:1.1,9.882332020385068)
--(axis cs:1.2,10.779882838459963)
--(axis cs:1.3,11.674245252314497)
--(axis cs:1.4,12.645126919851464)
--(axis cs:1.5,13.63036884984966)
--(axis cs:1.5,14.448403779792603)
--(axis cs:1.5,14.448403779792603)
--(axis cs:1.4,13.41093361972776)
--(axis cs:1.3,12.391418089498627)
--(axis cs:1.2,11.450627722332014)
--(axis cs:1.1,10.511116078615494)
--(axis cs:1.0,9.588652412529244)
--(axis cs:0.9,8.609435364476271)
--(axis cs:0.8,7.642974594686344)
--(axis cs:0.7,6.675779328078265)
--(axis cs:0.6,5.739998434533932)
--(axis cs:0.5,4.66814420865865)
--(axis cs:0.4,3.8490532393536725)
--(axis cs:0.3,2.6772265703021185)
--(axis cs:0.2,2.320474846545018)
--(axis cs:0.1,1.0578985630633322)
--cycle;

\path [fill=color2, fill opacity=0.2]
(axis cs:0.1,1.0056457956366107)
--(axis cs:0.1,0.9546896267354612)
--(axis cs:0.2,2.134236653591583)
--(axis cs:0.3,2.411617883371213)
--(axis cs:0.4,3.442676078773955)
--(axis cs:0.5,4.022902870996822)
--(axis cs:0.6,4.902153449417543)
--(axis cs:0.7,5.582991326467792)
--(axis cs:0.8,6.374255061900374)
--(axis cs:0.9,7.094444600172641)
--(axis cs:1.0,7.85906469426798)
--(axis cs:1.1,8.594380357216204)
--(axis cs:1.2,9.315732873546853)
--(axis cs:1.3,10.012253046086133)
--(axis cs:1.4,10.765584306846758)
--(axis cs:1.5,11.474105703978946)
--(axis cs:1.5,12.102863578573501)
--(axis cs:1.5,12.102863578573501)
--(axis cs:1.4,11.363110480767634)
--(axis cs:1.3,10.573830983831652)
--(axis cs:1.2,9.845469283410548)
--(axis cs:1.1,9.091466672991652)
--(axis cs:1.0,8.321603136370273)
--(axis cs:0.9,7.515938632864889)
--(axis cs:0.8,6.7593433738899)
--(axis cs:0.7,5.93090958543961)
--(axis cs:0.6,5.212604836256691)
--(axis cs:0.5,4.284558947918877)
--(axis cs:0.4,3.6720086067692446)
--(axis cs:0.3,2.5808752461237354)
--(axis cs:0.2,2.2907714642909545)
--(axis cs:0.1,1.0056457956366107)
--cycle;

\path [fill=color3, fill opacity=0.2]
(axis cs:0.1,0.6003611056873476)
--(axis cs:0.1,0.5697280689485632)
--(axis cs:0.2,1.2288209357808972)
--(axis cs:0.3,1.5307115143492813)
--(axis cs:0.4,2.1840470180667375)
--(axis cs:0.5,2.7103859571871785)
--(axis cs:0.6,3.3861667404962206)
--(axis cs:0.7,4.054419024799744)
--(axis cs:0.8,4.7543555369307775)
--(axis cs:0.9,5.453555342466271)
--(axis cs:1.0,6.148035140760893)
--(axis cs:1.1,6.845525138402846)
--(axis cs:1.2,7.569537919537203)
--(axis cs:1.3,8.28797340788661)
--(axis cs:1.4,9.02359040952562)
--(axis cs:1.5,9.79022126556337)
--(axis cs:1.5,10.349209968848049)
--(axis cs:1.5,10.349209968848049)
--(axis cs:1.4,9.542077698378487)
--(axis cs:1.3,8.769070672547286)
--(axis cs:1.2,8.017876130109782)
--(axis cs:1.1,7.256923518934044)
--(axis cs:1.0,6.526567905809303)
--(axis cs:0.9,5.792216461201808)
--(axis cs:0.8,5.049757509500637)
--(axis cs:0.7,4.309650231918669)
--(axis cs:0.6,3.5938706921914725)
--(axis cs:0.5,2.879161902425499)
--(axis cs:0.4,2.320277703191335)
--(axis cs:0.3,1.6271586421999122)
--(axis cs:0.2,1.3030093150529487)
--(axis cs:0.1,0.6003611056873476)
--cycle;

\addplot [semithick, color0, mark=diamond*, mark size=2, mark options={solid}]
table {%
0.1 1.030992248333661
0.2 2.246813956486008
0.3 2.592538730326873
0.4 3.727479716375554
0.5 4.520149588009402
0.6 5.561148988603966
0.7 6.468171665990052
0.8 7.407168042086496
0.9 8.34600757713244
1.0 9.297762867426025
1.1 10.19672404950028
1.2 11.115255280395989
1.3 12.032831670906562
1.4 13.028030269789612
1.5 14.039386314821131
};
\addplot [semithick, color2, mark=triangle*, mark size=2, mark options={solid}]
table {%
0.1 0.980167711186036
0.2 2.212504058941269
0.3 2.4962465647474743
0.4 3.5573423427715998
0.5 4.15373090945785
0.6 5.057379142837117
0.7 5.756950455953701
0.8 6.566799217895137
0.9 7.305191616518765
1.0 8.090333915319126
1.1 8.842923515103928
1.2 9.5806010784787
1.3 10.293042014958893
1.4 11.064347393807196
1.5 11.788484641276224
};
\addplot [semithick, color3, mark=*, mark size=2, mark options={solid}]
table {%
0.1 0.5850445873179554
0.2 1.265915125416923
0.3 1.5789350782745968
0.4 2.252162360629036
0.5 2.7947739298063388
0.6 3.4900187163438465
0.7 4.182034628359206
0.8 4.902056523215707
0.9 5.62288590183404
1.0 6.337301523285098
1.1 7.051224328668445
1.2 7.793707024823493
1.3 8.528522040216949
1.4 9.282834053952053
1.5 10.06971561720571
};
\end{axis}

\end{tikzpicture}\\
        \label{fig:is}
    \end{subfigure}
    \begin{subfigure}[h]{0.28\textwidth}
        \centering
        \begin{tikzpicture}

\definecolor{color0}{rgb}{0.12156862745098,0.466666666666667,0.705882352941177}
\definecolor{color1}{rgb}{1,0.498039215686275,0.0549019607843137}
\definecolor{color2}{rgb}{0.172549019607843,0.627450980392157,0.172549019607843}
\definecolor{color3}{rgb}{0.83921568627451,0.152941176470588,0.156862745098039}

\begin{axis}[
width = 5.5cm,
tick align=outside,
tick pos=left,
x grid style={white!69.0196078431373!black},
xlabel={\footnotesize Prediction Time (s)},
xmin=0.1, xmax=1.5,
xtick style={color=black},
y grid style={white!69.0196078431373!black},
ylabel={\footnotesize Dynamic MSE},
ymin=0, ymax=0.00377132888970664,
ytick style={color=black}
]
\path [fill=color0, fill opacity=0.2]
(axis cs:0.1,0.000409279062)
--(axis cs:0.1,0.000399646338)
--(axis cs:0.2,0.0009956528417000001)
--(axis cs:0.3,0.00112907316)
--(axis cs:0.4,0.001537540661)
--(axis cs:0.5,0.001724886841)
--(axis cs:0.6,0.002018951111)
--(axis cs:0.7,0.0021553433850000003)
--(axis cs:0.8,0.0023404338965)
--(axis cs:0.9,0.002436833734)
--(axis cs:1.0,0.0025886861030000002)
--(axis cs:1.1,0.002675526363)
--(axis cs:1.2,0.002774696524)
--(axis cs:1.3,0.0028714177420000003)
--(axis cs:1.4,0.002961460661)
--(axis cs:1.5,0.003020929137)
--(axis cs:1.5,0.003047274063)
--(axis cs:1.5,0.003047274063)
--(axis cs:1.4,0.002987546139)
--(axis cs:1.3,0.002897105858)
--(axis cs:1.2,0.002799950276)
--(axis cs:1.1,0.002700326837)
--(axis cs:1.0,0.002613082897)
--(axis cs:0.9,0.0024605074659999997)
--(axis cs:0.8,0.0023636369034999998)
--(axis cs:0.7,0.002177614015)
--(axis cs:0.6,0.002040508889)
--(axis cs:0.5,0.0017448201589999998)
--(axis cs:0.4,0.0015563657390000002)
--(axis cs:0.3,0.0011452172400000001)
--(axis cs:0.2,0.0010108181583)
--(axis cs:0.1,0.000409279062)
--cycle;

\path [fill=color2, fill opacity=0.2]
(axis cs:0.1,0.0004832396506)
--(axis cs:0.1,0.0004727696694)
--(axis cs:0.2,0.0011479251590000002)
--(axis cs:0.3,0.0012529260109999999)
--(axis cs:0.4,0.001648228899)
--(axis cs:0.5,0.001816960859)
--(axis cs:0.6,0.002078310647)
--(axis cs:0.7,0.002227588404)
--(axis cs:0.8,0.002413256761)
--(axis cs:0.9,0.002526205021)
--(axis cs:1.0,0.002694188043)
--(axis cs:1.1,0.0028019571649999997)
--(axis cs:1.2,0.002904914144)
--(axis cs:1.3,0.003017472269)
--(axis cs:1.4,0.003122051831)
--(axis cs:1.5,0.003200969245)
--(axis cs:1.5,0.0032280813549999997)
--(axis cs:1.5,0.0032280813549999997)
--(axis cs:1.4,0.003148828969)
--(axis cs:1.3,0.0030437995310000003)
--(axis cs:1.2,0.002930747856)
--(axis cs:1.1,0.002827330835)
--(axis cs:1.0,0.0027190713570000003)
--(axis cs:0.9,0.0025503035789999997)
--(axis cs:0.8,0.002436813239)
--(axis cs:0.7,0.002250224996)
--(axis cs:0.6,0.002100179353)
--(axis cs:0.5,0.0018374151409999999)
--(axis cs:0.4,0.001667715301)
--(axis cs:0.3,0.001269927189)
--(axis cs:0.2,0.001164201841)
--(axis cs:0.1,0.0004832396506)
--cycle;

\path [fill=color3, fill opacity=0.2]
(axis cs:0.1,0.0003035209314)
--(axis cs:0.1,0.0002952321686)
--(axis cs:0.2,0.000609508402)
--(axis cs:0.3,0.0008092114515)
--(axis cs:0.4,0.0010791227460000001)
--(axis cs:0.5,0.001304421554)
--(axis cs:0.6,0.0015185790975)
--(axis cs:0.7,0.0016895935765000002)
--(axis cs:0.8,0.001850938723)
--(axis cs:0.9,0.001980318758)
--(axis cs:1.0,0.002132957238)
--(axis cs:1.1,0.0022500250919999997)
--(axis cs:1.2,0.002367026375)
--(axis cs:1.3,0.0024833825120000002)
--(axis cs:1.4,0.00259788574)
--(axis cs:1.5,0.002689147065)
--(axis cs:1.5,0.0027140103349999997)
--(axis cs:1.5,0.0027140103349999997)
--(axis cs:1.4,0.00262232586)
--(axis cs:1.3,0.002507280488)
--(axis cs:1.2,0.002390360625)
--(axis cs:1.1,0.002272778108)
--(axis cs:1.0,0.002155113162)
--(axis cs:0.9,0.0020016712420000003)
--(axis cs:0.8,0.0018715848769999999)
--(axis cs:0.7,0.0017093238235)
--(axis cs:0.6,0.0015372891024999999)
--(axis cs:0.5,0.001321768446)
--(axis cs:0.4,0.001094907854)
--(axis cs:0.3,0.0008228909485)
--(axis cs:0.2,0.000621389198)
--(axis cs:0.1,0.0003035209314)
--cycle;

\addplot [semithick, color0, mark=diamond*, mark size=2, mark options={solid}]
table {%
0.1 0.0004044627
0.2 0.0010032355
0.3 0.0011371452
0.4 0.0015469532
0.5 0.0017348535
0.6 0.00202973
0.7 0.0021664787
0.8 0.0023520354
0.9 0.0024486706
1.0 0.0026008845
1.1 0.0026879266
1.2 0.0027873234
1.3 0.0028842618
1.4 0.0029745034
1.5 0.0030341016
};
\addplot [semithick, color2, mark=triangle*, mark size=2, mark options={solid}]
table {%
0.1 0.00047800466
0.2 0.0011560635
0.3 0.0012614266
0.4 0.0016579721
0.5 0.001827188
0.6 0.002089245
0.7 0.0022389067
0.8 0.002425035
0.9 0.0025382543
1.0 0.0027066297
1.1 0.002814644
1.2 0.002917831
1.3 0.0030306359
1.4 0.0031354404
1.5 0.0032145253
};
\addplot [semithick, color3, mark=*, mark size=2, mark options={solid}]
table {%
0.1 0.00029937655
0.2 0.0006154488
0.3 0.0008160512
0.4 0.0010870153
0.5 0.001313095
0.6 0.0015279341
0.7 0.0016994587
0.8 0.0018612618
0.9 0.001990995
1.0 0.0021440352
1.1 0.0022614016
1.2 0.0023786935
1.3 0.0024953315
1.4 0.0026101058
1.5 0.0027015787
};
\end{axis}
\end{tikzpicture}\\
        \label{fig:mse_dyn}
    \end{subfigure}
    \begin{subfigure}[h]{0.1\textwidth}
        \begin{tikzpicture}
\definecolor{color0}{rgb}{0.12156862745098,0.466666666666667,0.705882352941177}
\definecolor{color1}{rgb}{1,0.498039215686275,0.0549019607843137}
\definecolor{color2}{rgb}{0.172549019607843,0.627450980392157,0.172549019607843}
\definecolor{color3}{rgb}{0.83921568627451,0.152941176470588,0.156862745098039}

\begin{customlegend}[legend columns=1,legend style={align=left,draw=white!80!black,column sep=2ex},
        legend entries={\footnotesize PredNet,
                        \footnotesize D-P,
                        \footnotesize Ours}]
        \addlegendimage{semithick, color0, mark=diamond*, mark size=2, mark options={solid}}
        \addlegendimage{semithick, color2, mark=triangle*, mark size=2, mark options={solid}}
        \addlegendimage{semithick, color3, mark=*, mark size=2, mark options={solid}}
        \end{customlegend}
\end{tikzpicture}
        \label{fig:legend}
    \end{subfigure}
    \vspace{-0.2cm}
    \caption{\small The MSE, IS, and dynamic MSE metrics on OGM predictions are evaluated at each time step in the prediction horizon. We note that the cell-wise standard error is too small to be visible in the MSE and the dynamic MSE plots. Lower is better.}
    \label{fig:quantitative_plots}
\vspace{-0.6cm}
\end{figure*}
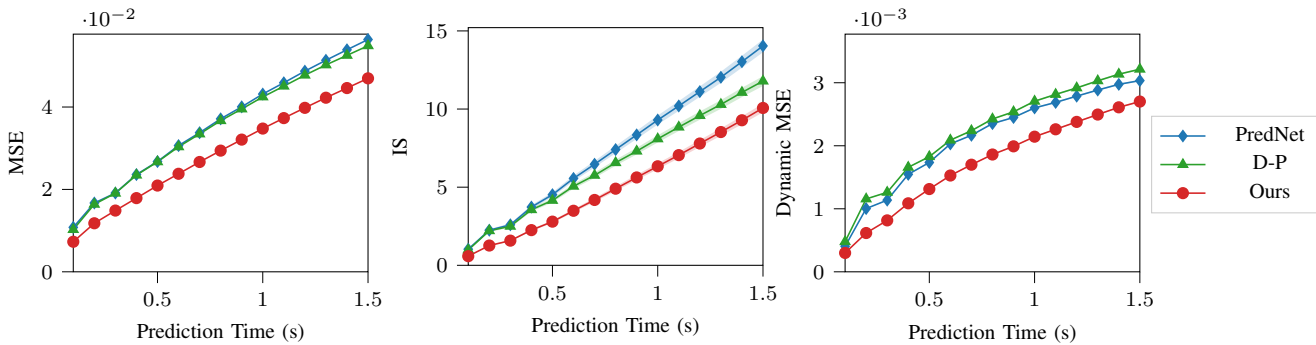
\subsection{Quantitative Evaluation}
\label{section:quant}
The mean squared error (MSE), and the image similarity (IS)~\cite{im_sim} metrics between the predicted OGMs and ground truth labels are calculated to evaluate the performance of our semantics-aware environment prediction model. The MSE metric is used to measure how accurately the predicted occupancy probability corresponds to the target value. The IS metric is a measure of how well the environment structure is retained in the predictions. Additionally, the dynamic MSE~\cite{mery_icra} is also determined to assess the prediction accuracy of occupancy probability of dynamic cells. 
To obtain the dynamic MSE value, we first apply the ground truth dynamic masks to both the predicted and ground truth OGM frames before computing the MSE.  
The ground truth dynamic masks are obtained by using object detection and tracking information provided in the Waymo Open Dataset~\cite{Sun_2020_CVPR}, following \citet{mery_icra}.

\cref{table:metrics} shows the MSE, IS, and dynamic MSE metrics averaged over the $15$ prediction time steps. The performance of our model is better than the baseline models for all metrics. Our model outperforms the double-prong~\cite{mery_iros} and PredNet~\cite{PredNet} models in MSE by $22.6$\%, and $25.1$\%, respectively.
\cref{fig:quantitative_plots} illustrates the plots of the MSE, IS, and dynamic MSE metrics versus the prediction time step for a \SI{1.5}{\second} prediction time horizon. 
Due to the accumulation of errors made in the predictions, all metric values increase with the prediction time step. This is expected since, for multiple-frame prediction, the models treat the predictions as actual inputs and cause the prediction errors to build up over time. 
Nevertheless, our model performs better in all prediction time steps across all metrics and is able to retain moving objects in the predictions for longer.
It is noted that the double-prong (D-P) model performs better than the PredNet model in both the MSE and IS metrics but slightly worse in the dynamic MSE, even though the double-prong model incorporates environment dynamics information.

In the original double-prong work, the training data is subsampled to contain only scenes with moving objects present in order to help with learning the dynamic OGMs in the dynamic prong. This is because of the data bias in the training data that mainly consist of static objects compared to moving objects. The results shown in \cref{table:metrics} are from models that are trained on the same training data without any subsampling for a fair comparison across all methods. We additionally train the double-prong model on subsampled data, and obtain the MSE ($\times 10^{-2}$), IS and dynamic MSE ($\times 10^{-3}$) metrics at $3.59 \pm 0.0011$, $7.00 \pm 0.066$, and $2.09 \pm 0.0024$, respectively. As expected, the double-prong model performs better in the dynamic MSE when trained on subsampled data. However, our method of using environment semantics still outperforms the double-prong model that uses environment dynamics across all metrics.

\subsection{Impact of Semantic Categories}
To study the influence of the types and number of semantic labels on occupancy prediction, we experimented with $3$ variants of SMGMs. The first consists of $4$ semantic labels that are vehicle (combining the car and other vehicles labels as described in \cref{section:exp}), cyclist, pedestrian, and \emph{others} labels, 
while the second variant contains $6$ semantic labels, with the addition of building and road labels since they make up the majority of the semantic data. The third variant is a binary SMGM that consists of vehicle and \emph{others} labels. The \emph{others} label consists of miscellaneous objects, along with the rest of the objects that are not considered in each variant. 

We obtain comparable performance for all $3$ variants as compared to the original SMGM data. The MSE ($\times 10^{-2}$), IS and dynamic MSE ($\times 10^{-3}$) metrics are $2.93 \pm 0.0010$, $5.14 \pm 0.051$, and $1.61 \pm 0.0020$, respectively, for the first SMGM variant, and $2.86 \pm 0.0010$, $5.32 \pm 0.049$, and $1.65 \pm 0.0021$, respectively, for the second variant, and $2.97 \pm 0.0010$, $5.52 \pm 0.054$, and $1.60 \pm 0.0020$, respectively, for the third variant. The results show that our model outperforms the baselines
across all metrics when using environment semantics for all SMGM variants and the original SMGM data. By using various movable object labels, the model is encouraged to learn different dynamics of specific moving objects. The static semantic labels has less impact on the accuracy of occupancy prediction (with respect to MSE and dynamic MSE metrics). However, using more static semantic labels helps to maintain the structure of the OGM predictions, as shown by a better IS metric in the original SMGM data compared to the variants.          

The improvement to the predictions is due to its semantics-aware nature. We hypothesize that the prediction accuracy increases because the model has access to both the occupancy probability and the object label of each cell and can deduce how the motion should be propagated spatially and temporally. Our method effectively and accurately predicts the occupancy states of the environment using relevant occupancy and valuable semantic information.    

\section{Conclusion}
Our semantics-aware spatiotemporal occupancy prediction framework 
successfully incorporates the environment semantic prediction module within the environment prediction model and improves prediction performance on real-world data. Compared to baseline methods, our proposed semantics-incorporated model has higher prediction accuracy and is able to retain the appearance of moving objects in the predictions for longer prediction time horizons.
Future work will extend the model to perform both semantic and occupancy prediction tasks concurrently, eliminating the need for separate modules and reducing the model size.

\section*{Acknowledgment}
We thank Bernard Lange for his help with implementing the image similarity metric calculation.

\renewcommand*{\bibfont}{\small}
\printbibliography

\end{document}